\documentclass[10pt,twocolumn,letterpaper]{article}

\usepackage[pagenumbers]{cvpr} 

\usepackage{graphicx}
\usepackage{subcaption}
\usepackage{adjustbox}
\usepackage{balance}
\usepackage{calc}
\usepackage{caption}

\makeatother

\definecolor{cvprblue}{rgb}{0.21,0.49,0.74}
\usepackage[pagebackref,breaklinks,colorlinks,allcolors=cvprblue]{hyperref}

\def\paperID{*****} 
\def\confName{CV4Animals}
\def\confYear{2025}

\title{Automated Detection of Salvin’s Albatrosses: Improving Deep Learning Tools for Aerial Wildlife Surveys}

\author{
\begin{tabular}{cccc}
Mitchell Rogers$^{1}$\thanks{Corresponding author: \texttt{mitchell.rogers@vuw.ac.nz}} &
Theo Thompson$^{2}$ &
Isla Duporge$^{3}$ &
Johannes Fischer$^{4}$ \\
Klemens Pütz$^{5}$ & Thomas Mattern$^{2,6,7}$ & Bing Xue$^{1}$ & Mengjie Zhang$^{1}$
\end{tabular}
\\
\\
$^1$Centre for Data Science and Artificial Intelligence, Victoria University of Wellington, New Zealand \\
$^2$Department of Zoology, University of Otago, New Zealand\\
$^3$Department of Ecology and Evolutionary Biology, Princeton University, U.S.A \\
$^4$Marine Bycatch and Threats – Department of Conservation, New Zealand\\
$^{5}$Antarctic Research Trust, Bremervörde, Germany\\
$^{6}$The Tawaki Trust, Dunedin, New Zealand \\
$^{7}$Global Penguin Society, Puerto Madryn, Chubut, Argentina
}

\begin{document}
\maketitle

\begin{abstract}
Recent advancements in deep learning and aerial imaging have transformed wildlife monitoring, enabling researchers to survey wildlife populations at unprecedented scales. Unmanned Aerial Vehicles (UAVs) provide a cost-effective means of capturing high-resolution imagery, particularly for monitoring densely populated seabird colonies. In this study, we assess the performance of a general-purpose avian detection model, BirdDetector, in estimating the breeding population of Salvin’s albatross (\textit{Thalassarche salvini}) on the Bounty Islands, New Zealand. Using drone-derived imagery, we evaluate the model's effectiveness in both zero-shot and fine-tuned settings, incorporating enhanced inference techniques and stronger augmentation methods. Our findings indicate that while applying the model in a zero-shot setting offers a strong baseline, fine-tuning with annotations from the target domain and stronger image augmentation leads to marked improvements in detection accuracy. These results highlight the potential of leveraging pre-trained deep-learning models for species-specific monitoring in remote and challenging environments.
\end{abstract}

\section{Introduction}
\label{sec:intro}
Advances in machine learning and aerial imaging techniques have greatly enhanced the scale and efficiency at which ecological data can be analysed \cite{Tuia_2022}. One of the most significant applications is the automation of population counts, a traditionally time-consuming process \cite{Gray_2018, Eikelboom_2019}. Automating population counts not only improves efficiency but also frees up resources for other critical forms of data collection during fieldwork.

It is possible to use very high-resolution satellite imagery to monitor animal populations at large spatial scales, as demonstrated for African elephants \cite{Duporge_2021}, Migratory wildebeest \cite{Wu2023} and numerous species of whales \cite{Cubaynes_2019,Gray_2019}, amongst others. However, smaller species that cannot be individually identified in satellite imagery, such as penguins and other seabirds, often require indirect estimation methods. These include detecting environmental indicators like guano stains \cite{Barber-Meyer_2007, LaRue_2014} or counting white dots, which are inferred to be the nesting birds \cite{Bowler_2020, Fretwell_2017, Rogers2023}.

Unmanned Aerial Vehicles (UAVs) provide a valuable alternative to satellites for capturing high-resolution imagery of wildlife over large areas, offering a more flexible and cost-effective approach than traditional ground surveys while covering broader regions \cite{Kholiavchenko_2024b, duporge2024}. Although UAVs cannot match the vast spatial coverage of satellites, they provide superior detail, making them particularly useful for monitoring wildlife populations that are difficult to detect from space. 

Deep learning is essential in UAV surveys as it enables automated processing of the vast image datasets these surveys generate, facilitating timely and accurate assessments of populations \cite{Christin_2019,Corcoran_2021}. In addition, drone-based automated counts have been shown to surpass the accuracy of observer-based counts \cite{Hodgson_2018}. In the context of bird detection, the BirdDetector model \cite{Weinstein_2022}, a general-purpose detector trained on datasets of many bird species, has demonstrated strong performance in both zero-shot settings, where birds are identified in a new environment without additional training, and fine-tuned settings using small amounts of domain-specific data. This generalisability reduces the annotation burden, making it easier to adapt the model for new ecological monitoring tasks, and makes BirdDetector more suitable for small bird datasets than training other common object detection models from scratch, such as the YOLO family of models \cite{Ma_2024,Mpouziotas_2023} or R-CNN models \cite{Hong_2019,Song_2024}.

This study improves on the BirdDetector model \cite{Weinstein_2022} and builds upon previous drone-based seabird population counting efforts \cite{Parker_2020,Mattern_2021,Hayes_2021} to develop a high-precision object detection model for estimating dense breeding populations of Salvin’s albatross (\textit{Thalassarche salvini}). Endemic to New Zealand, Salvin’s albatross is classified as \textit{``Vulnerable"} by the IUCN \cite{Thalassarche_salvini} facing persistent threats such as incidental bycatch in commercial fisheries \cite{Abraham_2019,Thompson_2014}. While existing models like BirdDetector perform well after fine-tuning, their accuracy can be further enhanced through stronger data augmentation strategies and optimised inference techniques. Developing improved drone-based population counting methods is essential to provide accurate and timely assessments of population size and trends, to ultimately support more effective conservation efforts for this threatened species \cite{Phillips_2016}. This species presents a particularly strong use case for this method, as it inhabits remote islands that are difficult to access, limiting fieldwork opportunities. Accelerating data collection through automated approaches allows ecologists to focus on gathering other critical field data rather than conducting manual counts. The combination of UAVs and deep learning-based post-processing is especially valuable in these remote environments.

\begin{figure*}[tb]
    \centering
    \begin{subfigure}{0.441\textwidth}
        \includegraphics[trim={0 0.2cm 0 0},clip,width=\linewidth]{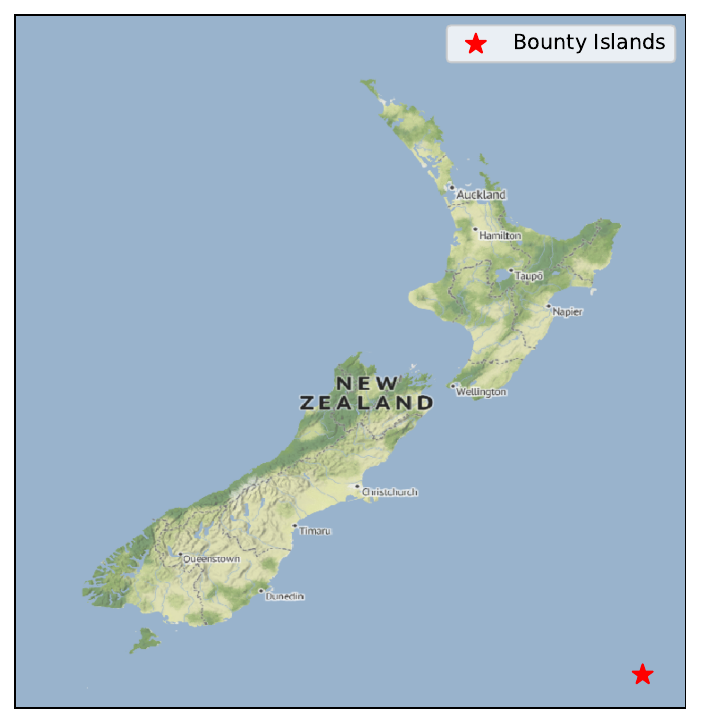}
        \caption{New Zealand and Bounty Islands.}
    \end{subfigure}
    \begin{subfigure}{0.529\textwidth}
        \includegraphics[trim={0 0.2cm 0 0},clip,width=\linewidth]{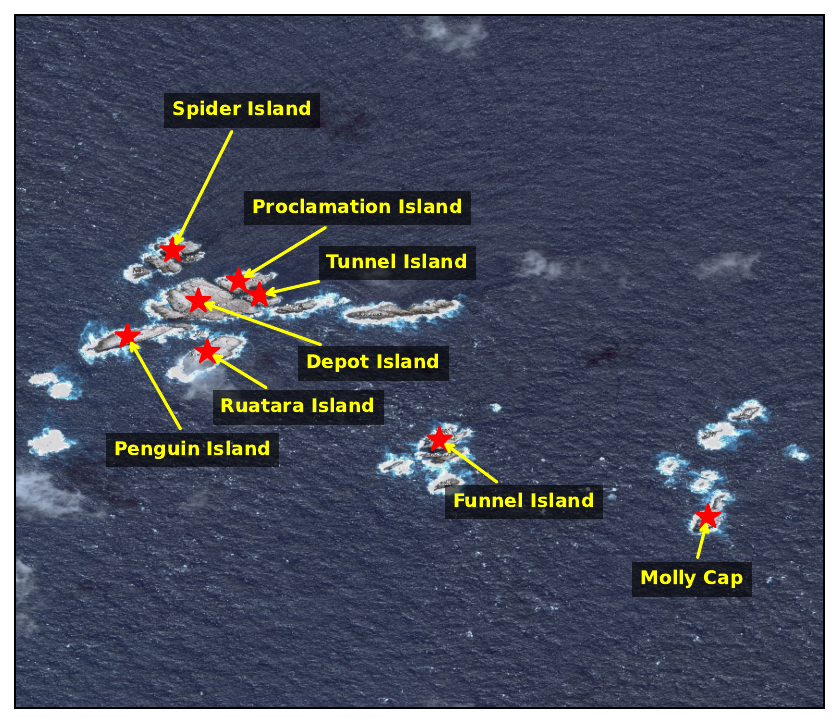}
        \caption{Zoomed-in Bounty Islands.}\label{fig:islands}
    \end{subfigure}
    \begin{subfigure}{0.96\textwidth}
        \includegraphics[trim={0 4cm 0 0},clip,width=\linewidth]{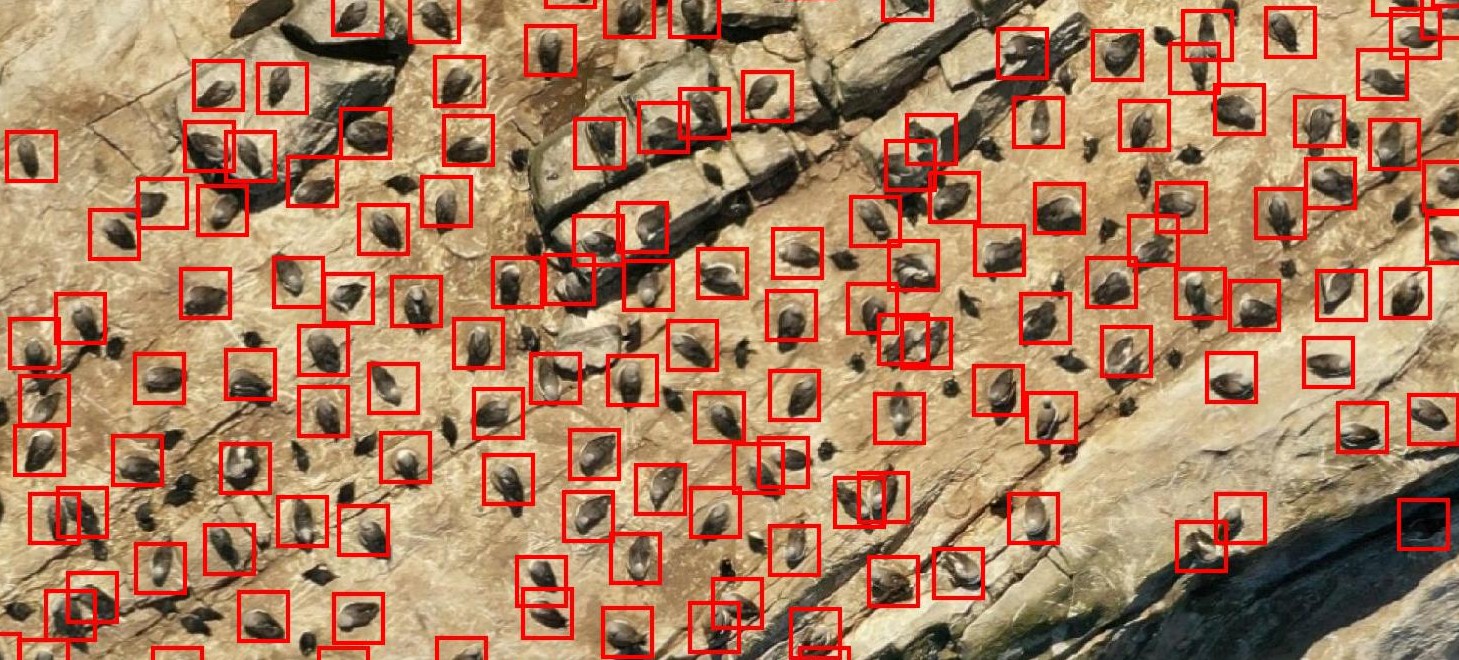}
        \caption{Example albatrosses observed from drone.}\label{fig:example}
    \end{subfigure}

    \caption{Location of the Bounty Islands relative to New Zealand, and example bounding boxes of the Salvin's Albatross.}
    \label{fig:maps}
\end{figure*}

\section{Data Overview}
This study focuses on eight granite islets in the Bounty Islands archipelago: Depot, Funnel, MollyCap, Penguin, Proclamation, Ruatara, Spider, and Tunnel Islands. The locations of these islands are shown in \cref{fig:islands}. Orthomosaic drone images of the Bounty Islands were collected by the Tawaki project \cite{Mattern_2023}, accompanied by manual point annotations for Salvin's albatrosses provided by multiple different annotators. For more details about the data collection process, see the previous work by Mattern et al. \cite{Mattern_2021}. Different from this previous study, a single set of annotations was selected per island by visually identifying the set with the most comprehensive coverage (i.e., the fewest missed albatrosses). While individual albatrosses were rarely missed due to their size, some parts of the island images were inadvertently missed by some annotators. Random tiles were also checked for errors. 

Point-based annotations were used due to their lower annotation effort compared to segmentation masks or bounding boxes, and their alignment with standard workflows of ecologists, in which individuals are manually counted using point annotations in image-based surveys. However, as observed in previous studies, inter-observer variability introduced some degree of noise \cite{Bowler_2020}. Additionally, annotations were occasionally offset from the centre of the individuals, reflecting the variability associated with manual labelling. These annotations were not originally designed for training deep learning models, and producing more precise, model-ready bounding-box annotations at this scale would require extensive manual effort spanning multiple days \cite{Parker_2020}.

To adapt these point annotations for object detection, pseudo-bounding boxes were generated by centring a 50~×~50-pixel square on each annotated point. This box size was selected to match the approximate size of albatrosses in the imagery. \Cref{fig:example} illustrates examples of these bounding boxes. While most ground-truth boxes closely align with individual birds, minor deviations occur in some cases. Each orthomosaic image was divided into non-overlapping supertiles measuring 1500~×~1500 pixels. The non-overlapping tiling strategy was employed to prevent data leakage between training and test sets. Only supertiles containing at least one annotation were retained, resulting in a final dataset of 571 tiles encompassing 66,635 annotated birds. 

\section{Models and Analysis}

We used the BirdDetector model \cite{Weinstein_2022} as the detection model. BirdDetector is adapted from the DeepForest model \cite{Weinstein_2019}, originally developed for detecting tree crowns in aerial imagery. Both models are built on RetinaNet \cite{RetinaNet} with a ResNet-50 backbone \cite{ResNet}. RetinaNet incorporates focal loss \cite{RetinaNet} to assign greater importance to difficult examples during training, thereby reducing the model overfitting to easily classified samples. Unless otherwise specified, the parameters used were the same as those used in the original BirdDetector paper. Two changes were made to this approach: inference was conducted using the Slicing-Aided Hyper-Inference (SAHI) method \cite{Akyon_2022}, and stronger augmentations were applied during training to improve the performance of the model on unseen images. SAHI aims to improve the performance of object detection models for small-object detection in high-resolution images. 


\subsection{Data Augmentation}

A key challenge in training object detection models for aerial imagery is handling variations in UAV flight altitude, which directly affect the apparent size of individuals in the images. This issue becomes particularly problematic when the test dataset contains images captured at different altitudes than those in the training set. BirdDetector addresses this challenge by applying augmentation techniques during training to simulate changes in object size and improve model generalisability \cite{Weinstein_2022}. In addition to variations in scale, background lighting and colour differences across images—caused by factors such as time of day and weather conditions—can introduce further challenges for model generalisability. In the original BirdDetector pipeline, data augmentation included random horizontal flipping, random cropping of 1000~×~1000 pixel regions, and random adjustments to brightness and contrast.

To enhance model generalisation, a more comprehensive augmentation strategy was implemented in this study. This revised approach incorporated additional transformations, including random modifications to brightness, contrast, and HSV (hue, saturation, and value) image values, as well as random flipping, and cropping and scaling operations. Specifically, the improved cropping strategy selected a single bounding box within the image and cropped a rectangular region with a randomly chosen width between 700 and 1200 pixels and an aspect ratio between 0.8 and 1.2. The cropped region was then resized to a standardised input size of 1000~×~1000 pixels. This cropping operation was applied to training images with a probability of 0.8.

\begin{table*}[tb]
    \centering
    \resizebox{\textwidth}{!}{
    \begin{tabular}{l|cccccccc|c}
       \toprule
       & \multicolumn{8}{c}{\textbf{Island}}\\
       \textbf{Model} & Depot & Funnel & MollyCap & Penguin & Proclamation & Ruatara & Spider & Tunnel & \textbf{Average}\\
       \midrule
        Zero-shot (-SAHI)           & 0.4828 & 0.5192 & 0.5319 & 0.5335 & 0.4824 & 0.5652 & 0.3752 & 0.5241 & 0.5018\\
        Zero-shot                   & 0.6983 & 0.7349 & 0.7215 & 0.6048 & 0.6297 & 0.696 & 0.506 & 0.6699 & 0.6576 \\
        Fine-tuned                  & 0.7548 & 0.7867 & 0.8066 & 0.6245 & 0.6855 & 0.7123 & 0.6247 & 0.6412 & 0.7045 \\
        Fine-tuned (+Augmentations) & \textbf{0.777} & \textbf{0.8244} & \textbf{0.8093} & \textbf{0.6678} & \textbf{0.754} & \textbf{0.7616} & \textbf{0.6421} & \textbf{0.767} & \textbf{0.7504} \\
       \bottomrule
    \end{tabular}}
    \caption{Overall F1 Score for each model variant across the eight island datasets.}
    \label{tab:Results_table}
\end{table*}

\subsection{Experimental Design}


Starting with the initial general BirdDetector model weights, four experiments were conducted. First, the pre-trained BirdDetector model was applied in a zero-shot setting without fine-tuning to benchmark the performance of the fine-tuned models, with and without SAHI for test inference. Second, the model was fine-tuned for 30 epochs using an initial learning rate of 0.0001 and the same augmentations and parameters as those of the original BirdDetector model. Third, the improved augmentation set was used, and the second experiment was repeated. For the fine-tuning experiments, 12.5\% of the training images per island were used to measure validation performance. The detection performance was measured in terms of the F1 score, using Leave-One-Island-Out Cross-Validation (LOIOCV), where each model was trained on data from seven islands and evaluated on the held-out island to estimate the generalisation performance on the unseen island. The number of images per island varied between 34 for Tunnel Island and 189 for Depot Island. Rather than subsampling an equal number of supertiles per island like other albatross counting studies \cite{Bowler_2020}, all tiles from the islands were used to simulate the effect of applying the trained models to a newly collected set of island drone images, and for transparency, the cross-validation F1 score per held-out island is presented.

\begin{figure*}[htb]
    \centering
    \begin{subfigure}[t]{0.49\textwidth}
        \includegraphics[width=\textwidth]{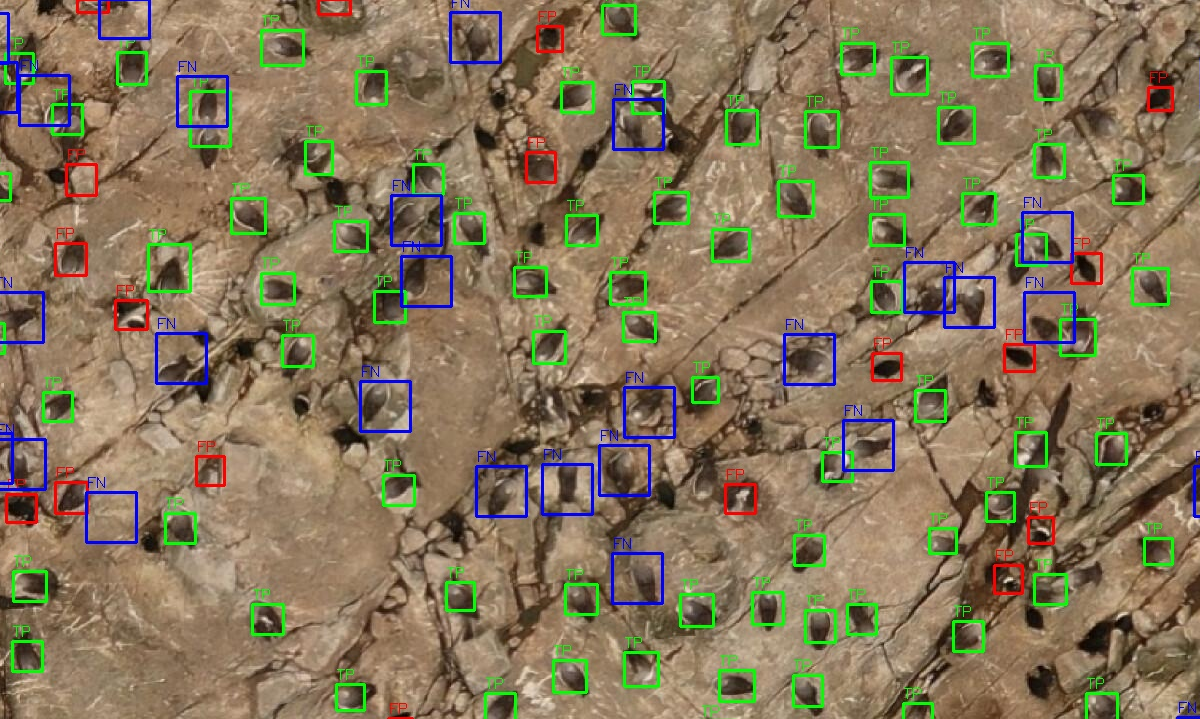}
        \caption{Zero-shot detections.}
        \label{fig:first}
    \end{subfigure}
    \hfill
    \begin{subfigure}[t]{0.49\textwidth}
        \includegraphics[width=\textwidth]{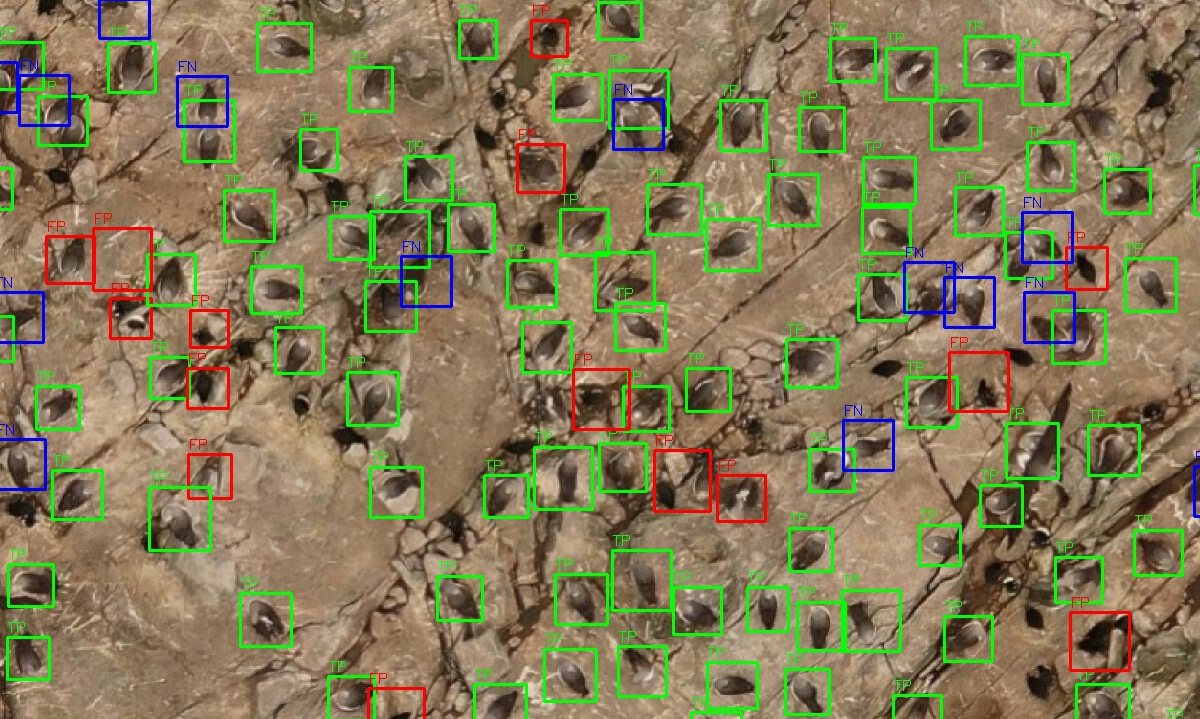}
        \caption{Improved fine-tuned model detections.}
        \label{fig:second}
    \end{subfigure}
    \caption{Example predictions for the zero-shot and improved fine-tuned experiments. The coloured boxes correspond to the true positives (green), false positives (red), and false negatives (blue), respectively.}
    \label{fig:example_images}
\end{figure*}

For evaluation, the intersection-over-union threshold, which defines the minimum acceptable intersection area between the true and predicted bounding boxes, was set to 0.1. This was set very low to account for possible inaccurate boxes generated from the annotated points. A low confidence score threshold (0.1) was used as many of the predicted boxes were assigned a low score, and a non-maximum suppression threshold of 0.05 was used. These are fixed thresholds, as with the DeepForest package \cite{Weinstein_2019}, used across all experiments, including the zero-shot experiments, and were not optimised using a validation set. For inference on the validation and test sets, tiling was performed using SAHI, where the images were sliced into 1000 × 1000-pixel overlapping patches, and the overlapping detections were merged back into the original image size. The zero-shot method was compared using SAHI for reference and based on resizing the 1500 × 1500-pixel tiles. Although this is not expected to make a significant difference for these small tiles, SAHI is expected to better handle detections in overlapping areas for full ortho-mosaiced images.

\section{Results}
\Cref{tab:Results_table} shows the F1 scores per island and experiment. The zero-shot results reflect the performance of the pre-trained model alone, while the others show results after fine-tuning on seven islands and testing on the eighth. The zero-shot model performed substantially better using SAHI for inference, suggesting that tile resolution influences the reliability of the predictions. However, these results do not demonstrate how this inference strategy manages predictions in overlapping tile regions, which can result in double counting if not addressed properly \cite{Brack_2018}. It is expected that the SAHI post-processing logic for merging predictions in overlapping tiles will reduce the number of double counts in large orthomosaic images compared to a na\"ive tiling approach \cite{Akyon_2022}. For future whole-island inference, applying the trained models to full orthomosaics using overlapping tiles in SAHI should improve the accuracy of population counts.

Despite performing reasonably well without training on the target domain, the zero-shot model produced significantly more false positives, with background objects, such as rocks, penguins and seals, being detected, and many albatrosses were missed. While penguins were not included in the current set of annotations, future work will aim to detect multiple species of birds.

As expected, the fine-tuned models performed markedly better overall than the zero-shot models. For some of the islands, fine-tuning using the normal augmentation strategy provided only a small improvement over the zero-shot model, and even in the case of Tunnel Island, the fine-tuned model performed worse. This was due to lower precision and more false positives from the fine-tuned model.

On most islands, the models trained with stronger augmentations achieved higher precision at the expense of a lower recall. Some islands posed greater challenges than others—for example, the strong contrast between rocks and shadows on Spider and Penguin Islands led to the frequent misidentification of terrain features as birds. In contrast, the performance was highest on the Funnel Island images, where shadows were less pronounced, and the albatross contrasted more clearly from the background. These results suggest that performance on new images may depend on their visual similarity to the pre-training data. \Cref{fig:example_images} compares the predictions of the model under zero-shot (with SAHI) and fine-tuned (with augmentations) settings for a portion of the Funnel Island. In both cases, the models struggled to separate overlapping bounding boxes and mistakenly detected multiple penguins.

By introducing greater variability during training, the stronger data augmentations help the model generalise better to new environmental conditions and reduce overfitting to features specific to the training samples.

\section{Discussions and Conclusions}

Overall, the improved fine-tuned models demonstrated strong performance, particularly when compared with both the zero-shot and baseline fine-tuned methods. When averaged over the eight islands, an average of 29.25\% of the detections were false positives, and only 19.3\% of birds were missed. Although there is room for improvement, this approach can substantially accelerate the current manual population-counting approach being used. We also expect the performance to improve further by retraining on the full dataset rather than holding out islands or using other unannotated images.

This study is part of a broader effort to develop better machine learning methods to monitor populations of nesting seabird colonies on remote islands surrounding New Zealand, using either very high-resolution satellite images \cite{Rogers2023} or drone surveys. Future work will extend this method to also detect other birds living among these albatrosses, such as the Erect-crested penguins (\textit{Eudyptes sclateri}) and the Bounty Island shag (\textit{Leucocarbo ranfurlyi}) which are visibly smaller, less distinguishable from the background, and more elusive, often hiding in rock niches. The Erect-crested penguins are one of the four penguin species worldwide classified as ``Endangered" by the IUCN Redlist \cite{Eudyptes_sclateri}. These penguins are difficult to spot manually, and even the zero-shot BirdDetector model rarely detects them. Following other wildlife detection approaches \cite{Beery_2019,PyTorch_Wildlife,Ma_2024b}, we plan to first detect all birds as a single class, then apply a secondary classifier to distinguish the individual species and the background. This two-stage approach should help reduce false positives by filtering background objects more effectively.

An alternative to the object-detection approach followed in this study is a counting-based approach to directly estimate population survey counts. When combined with detection, these models can also generate object locations, thereby improving the interpretability of the results \cite{CountGD}. In future work, we aim to investigate the use of pretrained object counting methods such as CountGD \cite{CountGD} and animal-specific approaches such as HerdNet \cite{HerdNet}.

{
    \small
    \bibliographystyle{ieeenat_fullname}
    \bibliography{main}
}


\end{document}